\pdfoutput=1

\documentclass[11pt]{article}

\usepackage[final]{acl}

\usepackage{times}
\usepackage{latexsym}

\usepackage[T1]{fontenc}

\usepackage[utf8]{inputenc}

\usepackage{microtype}

\usepackage{inconsolata}

\usepackage{graphicx}

\usepackage{bm}
\usepackage{booktabs}
\usepackage{multirow}
\usepackage{comment}
\usepackage{arydshln}
\usepackage{xcolor}
\usepackage{colortbl}

\usepackage{amssymb}

\definecolor{myred}{RGB}{206, 50, 16}
\definecolor{mygreen}{RGB}{20, 157, 51}
\newcommand{\deltaneg}[1]{\scriptsize{\bf \textcolor{myred}{(-{#1})}}}
\newcommand{\deltapos}[1]{\scriptsize{\bf \textcolor{mygreen}{(+{#1})}}}

\newcommand{\deltaneggreen}[1]{\scriptsize{\bf \textcolor{mygreen}{(-{#1})}}}
\newcommand{\deltaposred}[1]{\scriptsize{\bf \textcolor{myred}{(+{#1})}}}

\title{Vision-Language Models Struggle to Align Entities across Modalities}

\author{
  \textbf{Iñigo Alonso\textsuperscript{1}},
  \textbf{Gorka Azkune\textsuperscript{2}},
  \textbf{Ander Salaberria\textsuperscript{2}},
  \textbf{Jeremy Barnes\textsuperscript{2}},
  \textbf{Oier Lopez de Lacalle\textsuperscript{2}}
\\
\\
  \textsuperscript{1}School of Informatics, The University of Edinburgh \\
  \textsuperscript{2}HiTZ Center - Ixa, University of the Basque Country UPV/EHU \\
  \small{
    \textbf{Correspondence:} \href{mailto:ialonso@ed.ad.uk}{ialonso@ed.ad.uk}
  }
}

\begin{document}
\maketitle
\begin{abstract}
Cross-modal entity linking refers to the ability to align entities and their attributes across different modalities. While cross-modal entity linking is a fundamental skill needed for real-world applications such as multimodal code generation, fake news detection, or scene understanding, it has not been thoroughly studied in the literature. In this paper, we introduce a new task and benchmark to address this gap. Our benchmark, MATE, consists of 5.5k evaluation instances featuring visual scenes aligned with their textual representations. To evaluate cross-modal entity linking performance, we design a question-answering task that involves retrieving one attribute of an object in one modality based on a unique attribute of that object in another modality. We evaluate state-of-the-art Vision-Language Models (VLMs) and humans on this task, and find that VLMs struggle significantly compared to humans, particularly as the number of objects in the scene increases. Our analysis also shows that, while chain-of-thought prompting can improve VLM performance, models remain far from achieving human-level proficiency. These findings highlight the need for further research in cross-modal entity linking and show that MATE\footnote{Our dataset, evaluation results, and code are publicly available at \url{https://github.com/hitz-zentroa/MATE}} is a strong benchmark to support that progress. 
\end{abstract}

\section{Introduction}
\label{sec:intro}

Several real-world applications demand the ability to perform cross-modal entity linking, i.e., being able to align entities and attributes across modalities. In autonomous driving, for example, a single image of a scene may contain multiple entities, such as pedestrians and other vehicles. Additionally, textual or structured data about these entities, provided by smart devices or other cars, can include information like speed or future trajectory. While some attributes, such as vehicle color or shape, are shared between visual and textual representations, others, like speed, exist only in the text. To navigate effectively, the car must link vehicles in the image to their corresponding textual data, creating a unified representation of the scene. The same is true for other tasks such as multimodal code generation \cite{murobocodex, li2024mmcode}, multimodal fake news detection \cite{jing2023multimodal, ma2024event} and multimodal scene understanding \cite{su2024multimodal, li2024overview}.

\begin{figure}[t]
\centering
\includegraphics[scale=.27]{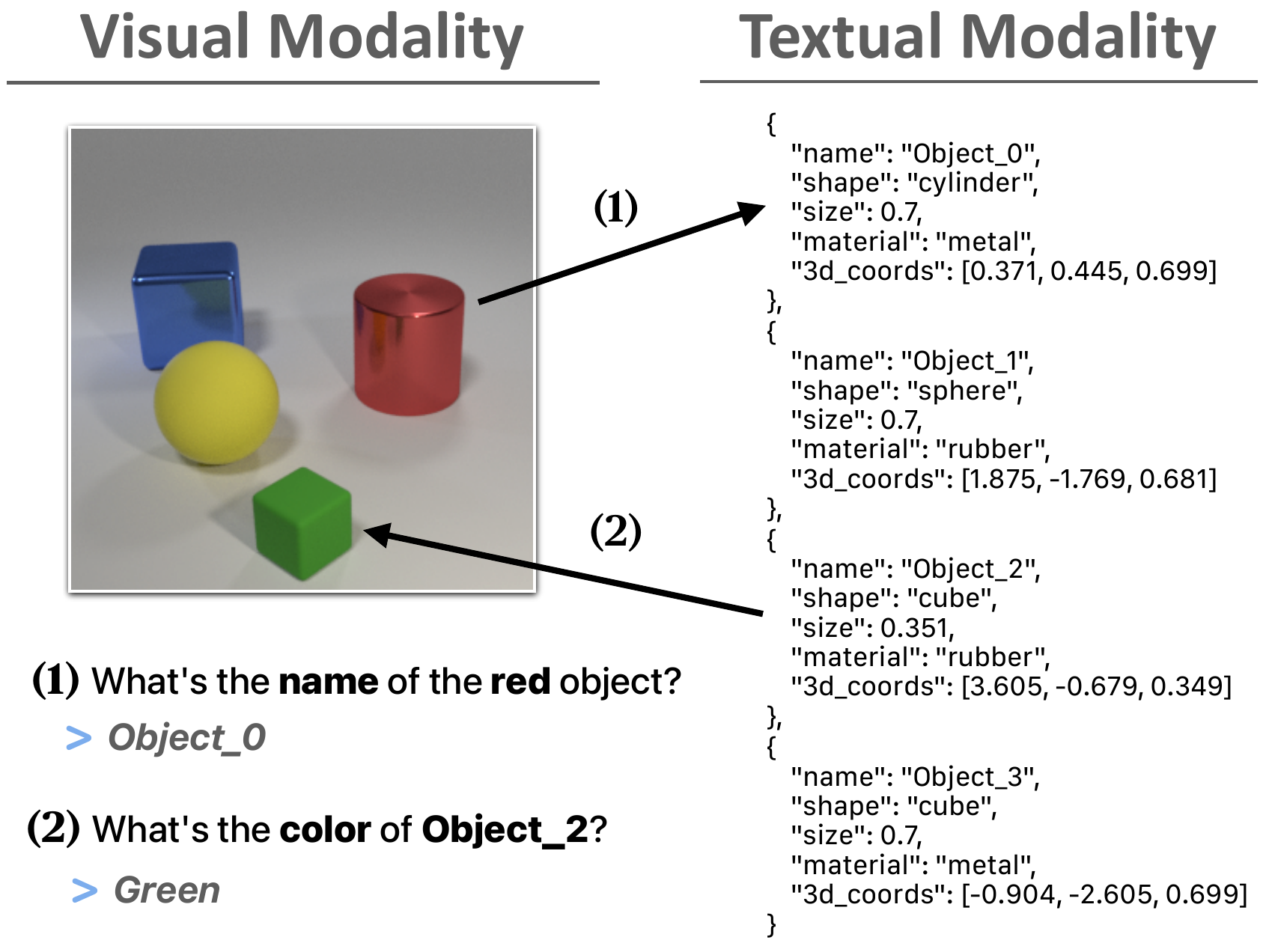}
\caption{Example of two MATE questions: (1)~\textbf{Image-to-text}: The model must identify the object in the image based on a visual attribute (red) and retrieve its name from the text ("Object\_0"). (2)~\textbf{Text-to-image}: The model must locate the object in the text using its name and determine its color, which is only present in the image. Both tasks require linking entities across modalities, but in opposite directions.}
\label{fig:overview}
\end{figure}

We can state, thus, that cross-modal entity linking is a basic ability needed to enable further applications of multimodal artificial intelligence systems. However, to the best of our knowledge, no exhaustive and targeted studies can be found in the literature. To fill that gap, in this paper we analyze the capabilities of current Vision-Language Models (VLM) for cross-modal entity linking. Specifically, we build a new multimodal question-answering benchmark called MATE that contains synthetic images of 3D scenes aligned with textual representations of those scenes (Figure \ref{fig:overview}). We use synthetic images as this methodology has a large advantage over real-world data -- namely, we can control for all the variables of the problem without interference from confounding factors such as object-attribute recognition or visual disambiguation, ensuring that we evaluate cross-modal entity linking independently. The task we use to evaluate the ability of models to perform cross-modal entity linking is shown in Figure \ref{fig:overview}. Given a pointer attribute which is unique in one of the modalities (e.g., the red color in the first question), we ask for a target attribute of that object which is shown in the other modality (e.g., the name of the referred entity). As the modalities we consider are visual and textual, we have image-to-text (question 1 of Figure \ref{fig:overview}) and text-to-image tasks (question 2), depending on where the pointer and target attributes exist.
%
We evaluate and analyze open- and closed-weight VLMs on MATE and perform further human evaluations. As a result of our experiments, we find that:

\paragraph{1) VLMs and humans show very different behaviors for cross-modal entity linking:} While humans achieve almost perfect performance, current VLMs fail to consistently align attributes across modalities out-of-the-box. Furthermore, VLMs are generally worse for the image-to-text variant; humans show balanced performance. Finally, the performance of VLMs is heavily influenced by the number of objects in the scene and the target attribute requested, whereas human performance is stable across different configurations.

\paragraph{2) The real challenge of our task resides in the cross-modal setting:} We show that VLMs can proficiently solve the task of entity-attribute linking when only one modality is considered. However, VLM performance decreases significantly as the number of attributes required to link entities across both modalities increases.

\paragraph{3) MATE is an effective benchmark for evaluating cross-modal entity linking in VLMs:} Our results indicate that MATE is a useful resource for evaluating the cross-modal alignment capabilities of current models. While human performance remains consistently high, VLMs suffer a significant drop as scene complexity increases, highlighting the challenges that cross-modal entity linking presents for current models.


\section{Related Work}
\label{sec:sota}
Several multimodal tasks in the literature are related to the one we propose, often grouped under the generic term of \textit{visual grounding}. For example, Referring Expression Comprehension (REC) \cite{kazemzadeh2014referitgame} requires identifying the image region described by a textual mention, typically referring to objects or physical entities along with their attributes and relations to other objects. Similarly, the Situated and Interactive Multimodal Conversations dataset (SIMMC) \cite{moon-etal-2020-situated,kottur-etal-2021-simmc} introduces a multimodal dialogue task where a system assists users in a virtual shopping scenario. To complete the task, the system must link visual objects to their textual metadata and search for relevant information. While both tasks share the challenge of aligning visual and textual content, our task extends this further by requiring explicit cross-referencing of object attributes across modalities. In particular, SIMMC avoids this challenge in its shared task by providing gold object IDs, eliminating the need for linking from raw multimodal inputs.

Multimodal Entity Linking (MEL) \cite{gan2021multimodal, adjali2020multimodal, song2024dual} is a related task where mentions in multiple modalities are disambiguated by linking them to the corresponding named entities in a knowledge base such as Wikipedia. While previous research \cite{yao-etal-2024-ameli} has primarily focused on scenarios where the image provides supporting visual information for a single entity, our cross-modal entity linking setup involves multiple entities present in both the image and the knowledge base. Similarly to MEL, our task also requires linking textual mentions with visual regions, but we go further by considering pointer attributes of entities (e.g., color and shape) and ensuring that both visual and textual descriptions of the objects are linked to accurately solve the task. This requires the model to perform both visual and textual searches to establish a coherent link across modalities.

All these tasks and many other related ones rely on the core ability of \textit{visual search}, the process of efficiently recognizing and localizing key objects within given scenes, a long-studied topic in cognitive sciences \cite{peelen2011neural, wolfe2020visual, wolfe2011visual}. Several computational models have been proposed for visual search, showing the difficulties of matching human performance \cite{sclar2020modeling, yang2020predicting, zhang2018finding, wu2024v}. Recent work  \cite{campbell2024understanding} proposes that those difficulties are closely related to the binding problem \cite{roskies1999binding}, i.e., the ability to associate one feature of an object (e.g., its color) with the other features of that object (e.g., its shape and location).

Our task is also based on visual search, but it adds the homologous textual search and poses the challenge of \textit{linking} the textual description of an object (with a given set of attributes) with its visual description using a unique pointer attribute (Section \ref{sec:methodology}). To the best of our knowledge, no similar task has been studied previously.

\section{Cross-modal Entity Linking}
\label{sec:methodology}

Cross-modal entity linking refers to the ability to understand that an entity described in two different modalities is actually the same entity. For example, in Figure \ref{fig:overview}, the red cylinder is represented in two ways: visually, with attributes such as color, shape, and size, and textually, as a set of attribute-value pairs. 

In order to evaluate whether VLMs possess this ability, we propose a question-answering task. We create 3D scenes containing multiple objects and provide a textual collection of attribute-value pairs for all objects in the scene (see Figure \ref{fig:overview}). Importantly, while some attributes are shared across modalities, others are exclusively available in one modality. In Figure \ref{fig:overview}, the color attribute is not included in the textual modality, while the object's name appears only in text and other attributes (shape, size, and material) are common to both.

Using this setup, we design questions that ask for the value of a particular object's attribute. In Figure \ref{fig:overview}, the first question asks for the name of the red object. In this example, the color red acts as the \textit{pointer attribute}, which identifies the object in the scene. To avoid ambiguity, each pointer attribute is unique to a single object. The “name” attribute, only available in the textual modality, serves as the \textit{target attribute}. Answering correctly requires linking the object's visual and textual representations.

Question 1 in Figure \ref{fig:overview} is an \textit{image-to-text} task because the pointer attribute is only represented in the visual modality and the target attribute in the text. Question 2, instead, is \textit{text-to-image}, where the pointer attribute “name” is only represented in text, and the target attribute is the color, represented only in the image.
This question-answering task is a suitable proxy for testing the ability of cross-modal entity linking, since to achieve high performance, models must link two different representations of the same entity.

\subsection{The MATE Benchmark} 
To perform this evaluation, we introduce MATE, a benchmark dataset consisting of 5,500 question-answering examples. Each example features a scene composed of three to ten 3D geometric objects with various colors, shapes, materials, and sizes (see Figure \ref{fig:overview} for reference). Each scene is represented in both the visual modality (image) and the textual one as a list of objects and their attributes (shown in JSON format in Figure \ref{fig:overview}). The scenes in MATE are based on the CLEVR dataset~\cite{johnson2017clevr}, but we extend them with additional shapes and uniquely identifiable object names.

MATE includes one question per example, and each question features a pointer and a target attribute. When the pointer or target attribute belongs to the visual modality, we use \textit{color} or \textit{shape}. For attributes residing in the textual modality, we use \textit{name}, \textit{rotation}, \textit{size}, and \textit{3D coordinates}. Additionally, the dataset features a \textit{material} attribute, which, although not used as a pointer or target due to its limited value range, still serves as a descriptive property (see Appendix \ref{app:attributes} for a list of all attributes). Note that even though every serialized scene in our dataset contains all these attributes, the scene included in the prompt never contains the attribute pointed to or retrieved from the image. This prevents models from relying solely on a single modality. For example, in Figure \ref{fig:overview}, because the color attribute acts as the pointer attribute in the image-to-text question and as a target attribute in the text-to-image question, it is never included in the serialization provided to the model.

To ensure unbiased evaluation, MATE maintains a balanced distribution of features. The dataset examples are uniformly distributed across the number of objects in the scene. It also provides equal numbers of examples for both image-to-text and text-to-image tasks, with a total of 2,750 examples per setting. Furthermore, pointer-target attribute pairs are uniformly distributed across all object counts, resulting in $43 \pm 1.5$ examples per attribute pair, object count, and setting. This balanced design makes MATE a robust benchmark for evaluating cross-modal entity linking (see Appendix \ref{app:dataset_format} for all information included in the dataset).

\begin{table}[t]
\begin{tabular}{llccc}
\toprule
 & Model & Img2Txt & Txt2Img & Avg.\\
\midrule
     & Human & 97.9 & 97.9 & 97.9\\
 & Random & 25.4 & 18.5 & 22.0\\
\midrule
\multirow{5}{*}{\rotatebox{90}{Open}} 
& LLaVA 1.5 & 29.3 & 35.7 & 32.5 \\ 
& LLaVA 1.6 & 48.7 & 61.6 & 55.2 \\
& Molmo & 18.1 & 20.9 & 19.5\\
& Llama 3.2 & 37.4 & 11.4 & 24.4\\
& Qwen2-VL & 72.1 & 77.2 & 74.7 \\
& Qwen2.5-VL & \textbf{75.7} & \textbf{84.5} & \textbf{80.1} \\
\midrule
\multirow{3}{*}{\rotatebox{90}{Closed}} & Gemini 1.5 &  63.2 & 71.2 & 67.2 \\
& GPT-4o & 76.4 & 79.1 & 77.8 \\
& Claude 3.5 &  \textbf{80.9} & \textbf{85.7} & \textbf{83.3} \\
\bottomrule
\end{tabular}
\caption{Results of open and closed VLMs in our task. All results are obtained using two-shot prompting. Exact match accuracy is provided for image-to-text (Img2Txt column) and text-to-image (Txt2Img column) configurations. Human and random accuracies are shown as reference.}
\label{tab:main_results}
\end{table}

\begin{figure}[t]

\includegraphics[width=\linewidth]{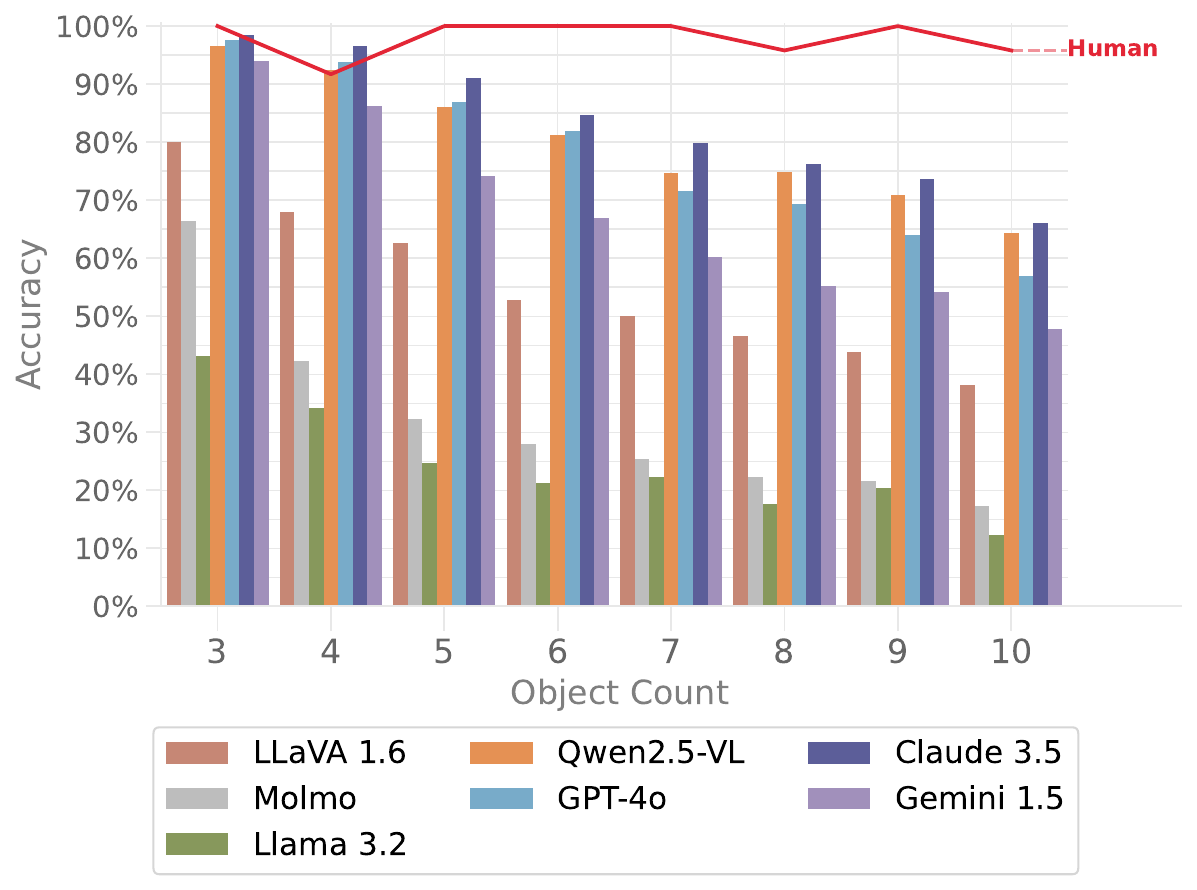}
\caption{Average accuracy of VLMs and humans for cross-modal entity linking, depending on the number of objects in the scene. VLM performance decreases with the number of objects, whereas humans perform very similarly for all scenarios.}
\label{fig:obj_count_performance}
\end{figure}

\section{Experiments}

\label{sec:experiments}
We conducted our experiments using a variety of open and closed models (see Appendix \ref{app:all_results} for a full list). For open models, we report results for the top-performing model from each family. The families we considered are LLaVA-1.5 \cite{liu2024visual}, LLaVA-1.6 \cite{liu2024llavanext}, Molmo \cite{deitke2024molmo}, Llama 3.2 \cite{dubey2024llama}, and Qwen2.5-VL \cite{Qwen2.5-VL}.\footnote{These VLMs can be paired with different LLM backbones; we only present results for the best-performing variants: LLaVA 1.5 13B, LLaVA 1.6 34B with Nous-Hermes-2-Yi-34B LLM backbone, Molmo-7B-O-0924, Llama-3.2-11B, Qwen2-VL-7B, Qwen2.5-VL-7B} For closed models, we included GPT-4o, Claude 3.5 and Gemini in our evaluation.\footnote{Tested versions: \textit{gpt-4o-2024-11-20}, \textit{claude-3-5-sonnet-20241022} and \textit{gemini-1.5-flash}} All open-weight models where evaluated in a computing cluster using four NVIDIA A100 80GB GPUs. The entire set of experiments was estimated to have required approximately 300 GPU hours. For the closed-weight commercial models, we used their respective official APIs.

We tested all models using zero-, one-, and two-shot prompting. As most evaluated models restrict visual input to a single image, we used one image along with two question-answer examples for two-shot prompting. When the target attribute resided in the visual modality, we included a set of possible options to help the models align their responses with the expected terminology (see Appendix \ref{app:croxss_prompt}). 

We represented the scene in the textual modality using JSON format. In an initial comparative study, we explored how different textual representation formats (JSON, YAML, XML, or verbalized descriptions) affected performance. The results showed no statistically significant differences between formats, though JSON and verbalized representations slightly outperformed the others. Thus, all experiments in this work were conducted using JSON as the scenes' representation format in the textual modality.

\subsection{Human Evaluation}
We furthermore conducted a human evaluation to quantify the extent to which humans are capable of performing the kind of entity-attribute correlations we expect from VLMs, which serves as an upper-bound for the task. 

For this purpose, we selected a subset of 384 examples from our benchmark, ensuring a representative distribution of features.\footnote{This subset included at least one combination of every target and pointer attribute across examples with all possible object counts from both image-to-text and text-to-image variants.} The evaluation examples were distributed among 5 participants (see Appendix \ref{app:participants} for further details). All of the evaluators were presented with the same image and textual prompts as those used in the VLM evaluations. 

\begin{table*}[th]
    \centering
\begin{tabular}{llrrrrrrr}
\toprule 
& & \multicolumn{2}{c}{{Txt2Img}} & \multicolumn{4}{c}{{Img2Txt}} & \multirow{2}{*}{\vspace{-0.2cm}OSE $\downarrow$} \\  \cmidrule(lr){3-4}  \cmidrule(lr){5-8} 
& Model & Shape & Color & Name & Coords & Rotation & Size & \\
\midrule
& Human & 100.0 & 93.2 & 96.9 & 96.9 & 96.9 & 100.0 & 0.0\\
& Random & 24.8 & 12.2 & 17.4 & 18.4 & 18.4 & 47.6 & N/A\\
\midrule
\multirow{5}{*}{Open} & LLaVA 1.5 & 47.5 & 24.0 & 27.8 & 13.7 & 24.4 & 51.5 & 18.1 \\
& LLaVA 1.6 & 70.4 & 53.0 & 56.0 & 35.3 & 37.1 & 66.5 & 6.5\\
& Llama 3.2 & 7.0 & 15.7 & 49.5 & 15.5 & 34.4 & 50.2 & 83.5\\
& Molmo & 31.4 & 10.5 & 21.8 & 32.9 & 4.8 & 12.9 & 47.1\\
& Qwen2-VL & 86.8 & 67.8 & 76.9 & 62.4 & 63.9 & 85.2 & \textbf{2.6}\\
& Qwen2.5-VL & \textbf{91.7} & \textbf{77.4} & \textbf{81.5} & \textbf{64.8} & \textbf{69.9} & \textbf{86.7} & 7.0\\
\midrule
\multirow{2}{*}{Closed} 
& Gemini 1.5 & 74.6 & 67.9 & 57.7 & 53.4 & 60.2 & 81.8 & \textbf{2.1}\\
& GPT-4o & 82.2 & 76.0 & 74.0 & 72.2 & 73.0 & 86.4 & 4.4\\
& Claude 3.5 & \textbf{90.6} & \textbf{80.3} & \textbf{78.8} & \textbf{78.4} & \textbf{75.1} & \textbf{91.2} & 3.6\\
\bottomrule
\end{tabular}
    \caption{Accuracy of VLMs for the text-to-image and image-to-text tasks, depending on the target attribute. Human and random performance are also depicted as reference. OSE, Out-of-Scene Error: the percentage of erroneous predictions that do not match any existing target attribute in the scene. Lower scores indicate that the model's errors are more likely due to entity linking mistakes rather than copying or hallucination issues.}
    \label{tab:pointer-target-attr}
\end{table*}

\section{Results}
\label{sec:results}
We report the exact match accuracy across all experiments, as they involved either extractive question-answering (for textual target attributes) or multiple-choice question-answering (for visual target attributes). We also report results for the two-shot settings due to their better performance and stability across inferences. We consider this setting a better representation of the capabilities of these models in this task (see Appendix \ref{app:shot_experiments}).

Table \ref{tab:main_results} presents the overall results of the VLMs on our MATE benchmark, divided into image-to-text and text-to-image variants, as well as their average. While the task is straightforward for humans (approaching 100\% accuracy), it remains challenging for current VLMs, although their performance is well above random chance. The best-performing open-weight VLM falls 17.8 absolute points behind human accuracy, highlighting significant room for improvement. Even the highest-performing closed-weight commercial VLM falls short of human performance by 14.6 absolute points. This gap is particularly notable given the expectation of near-perfect accuracy on such a fundamental task. Therefore, we can conclude that current VLMs, unlike humans, struggle to consistently link representations across different modalities for the same set of objects, which may limit the capacities of those models for more advanced tasks where this ability is required.

Nevertheless, the results reveal a clear progression in VLM capabilities. There is a noticeable trend between model performance, parameter count, and release date, with more recent and larger models consistently outperforming their counterparts (see complete results in Appendix \ref{app:all_results}). 

It is also interesting to note that the text-to-image configuration is easier for all VLMs,\footnote{The only exception is Llama3.2. We ignore the reasons behind this phenomenon, but we speculate it could be related to model architecture (the use of cross-attention layers to connect the visual encoder and LLM decoder) and training recipe (the LLM decoder is kept frozen during multimodal training).} indicating that it is easier to identify the pointer attribute in the text and link it with the visual attribute of the image. This is not the case of humans, who have the same performance for both task configurations. 

In Figure \ref{fig:obj_count_performance} we further analyze the performance of VLMs and humans as the number of objects in the scene increases (3-10). The performance is calculated as the average between image-to-text and text-to-image accuracies. When correlating attributes and linking entities correctly, the number of objects in a scene does increase the difficulty of the task, but humans show no degradation in their performance. On the contrary, for VLMs, performance degradation was significant, even for top-performing models like Claude, which is almost 30 absolute points behind humans for scenes of 10 obejcts. This behavior can be explained by \textit{feature interference}, which tends to increase with the number of objects. According to \cite{campbell2024understanding}, humans are more robust to feature interference than current VLMs, as observed again in our experiments.

Finally, Table \ref{tab:pointer-target-attr} shows the accuracies obtained by VLMs for each target attribute. In theory, once the object has been linked across modalities, copying an attribute as the answer should be equally straightforward for all attributes, especially in the image-to-text setting. However, our results indicate that VLM performance decreases as the range of possible values for the target attribute increases. Moreover, the low percentage of Out-of-Scene Errors reported in Table \ref{tab:pointer-target-attr} (percentage of erroneous predictions that do not match any existing target attribute value in the scene), suggests that incorrect predictions still correspond to other objects in the scene. This indicates that these errors are more likely caused by entity linking issues rather than copying or hallucination errors. The exceptions are Llama 3.2 and Molmo, with very high OSE, which accounts for their tendency to hallucinate attribute values.

\section{Analysis}
\label{sec:analysis}
Our human evaluations revealed that solving the cross-modal entity linking task typically involves a series of sequential steps. We use the image-to-text scenario in Figure \ref{fig:overview} (question 1) as an example:

\paragraph{1) Visual search:} The pointer attribute (the red color) is used to identify the object in the image (the red cylinder).

\paragraph{2) Linking attribute identification:} Attributes other than the pointer attribute, which distinguish the object from the others in the scene, are identified. In question 1 of Figure \ref{fig:overview}, the most distinctive linking attribute is the object's cylinder shape, as the red object is the only cylinder in the scene.\footnote{Our benchmark dataset includes all possible linking attribute combinations that uniquely identify the target entity in each example.}

\paragraph{3) Textual search:} The linking attributes identified in the previous step are used to locate the object in the textual modality. For instance, the object with the attribute \textit{"shape": "cylinder"} is found in the textual data, and its corresponding target attribute, \textit{"name"}, is retrieved.

For the text-to-image setting, the same steps are followed, but in reverse order.
In question 2 of Figure \ref{fig:overview}, for instance, step 2 would involve combining at least two attributes, such as shape and material, or shape and size. 
In our analysis, we break down the cross-modal entity linking task into these three subtasks to analyze VLM performance and identify the most challenging aspects for these models.

\subsection{Visual and textual search}
\label{sec:unimodal}
In this section, we evaluate whether VLMs can perform entity-attribute alignment within a single modality 
(steps 1 and 3 as identified by human annotators). This helps determine whether the poor multimodal results stem from a general failure to extract attributes within the same modality or from the challenges of cross-modal reasoning itself, i.e., linking attribute identification (step 2). 

We conduct question-answering tasks within the same modality for both the pointer and target attributes, which we call image-to-image (where both attributes exist in the visual modality) and text-to-text (where both attributes exist in the textual modality).\footnote{Due to implementation issues, the tested VLMs require an input image even for the text-to-text experiments. We ran all the experiments twice, using white images and black images as inputs and found that the performance did not vary. Therefore, we report results using a white image for the text-to-text task.} Following the terminology used by \citet{campbell2024understanding}, we are evaluating disjunctive visual and textual searching capabilities separately. 

For each case, models were prompted with the task's objective and provided two solved questions as a two-shot prompting setting. For image-to-image tasks, we included a set of possible answer options in order to help the models align their responses with the terminology of hidden attributes (see Appendix \ref{app:single_prompt} for examples of these unimodal prompts).

Table \ref{tab:unimodal_results} reports the results of the unimodal search experiments (image-to-image and text-to-text). All models show significantly stronger performance compared to the cross-modal entity linking (Table \ref{tab:main_results}). In particular, Qwen2-VL, Qwen2.5-VL, and the closed-weights commercial VLMs achieve near-perfect accuracy, although a number of VLMs still struggle with entity search. In a number of  cases, we observe certain performance discrepancies between image-to-image and text-to-text tasks. We hypothesize that, despite having strong visual capabilities, the LLM backbones of these VLMs play a critical role in determining overall performance, and thus text-to-text performance tends to be higher.

In Figure \ref{fig:obj_count_performance_unimodal} (see Appendix~\ref{app:3dc_lnk_attr_overlap}), we plot the performance of VLMs for different numbers of objects in the scenes, reporting the average accuracies for both image-to-image and text-to-text tasks. The results show that model performance is much more stable than in the cross-modal scenario (Figure \ref{fig:obj_count_performance}), indicating that the number of objects is not a significant factor when only a single modality is considered. These findings align with \cite{campbell2024understanding}, whose results on disjunctive visual search tasks show that the number of objects does not affect VLM performance when feature interference is low.

Qwen2.5-VL performs near perfectly in unimodal tasks; however, it struggles with cross-modal ones (Table \ref{tab:main_results}), suggesting that its difficulties in MATE 
stem from the second step: linking attribute identification, making it a perfect candidate for further analysis. In Section \ref{sec:linking-attributes}, we will examine the performance of this model in greater depth.

\begin{table}
\begin{tabular}{llccc }
\toprule
 & Model & Img2Img & Txt2Txt & Avg. \\
\midrule
 & Human & 100.0 & 99.0 & 99.5\\
 & Random & 18.7 & 25.3 & 22.0\\
\midrule
\multirow{5}{*}{\rotatebox{90}{Open}} 
& LLaVA 1.5 & 55.3 & 88.6 & 72.0\\
& LLaVA 1.6 & 84.4 & 98.0 & 91.2\\
& Molmo & 86.9 & 54.7 & 70.8\\
& Llama 3.2 & 68.9 & 97.1 & 83.0\\
& Qwen2-VL & \textbf{99.7} & 98.1 & 98.9\\
& Qwen2.5-VL & \textbf{99.7} & \textbf{99.4} & \textbf{99.5}\\
\midrule
\multirow{3}{*}{\rotatebox{90}{Closed}} 
& Gemini 1.5 & 95.9 & \textbf{100.0} & 98.0\\
& GPT-4o & \textbf{98.4} & \textbf{100.0} & \textbf{99.2}\\
& Claude 3.5 & 97.3 & \textbf{100.0} & 98.7\\
\bottomrule
\end{tabular}
\caption{Results of open and closed VLMs in the unimodal variants of our task. All results are obtained using two-shot prompting. Exact match accuracy is provided for image-to-image (Img2Img column) and text-to-text (Txt2Txt column) configurations. Human and random accuracies are shown as reference.}
\label{tab:unimodal_results}
\end{table}

\begin{figure}[t]

\includegraphics[width=\linewidth]{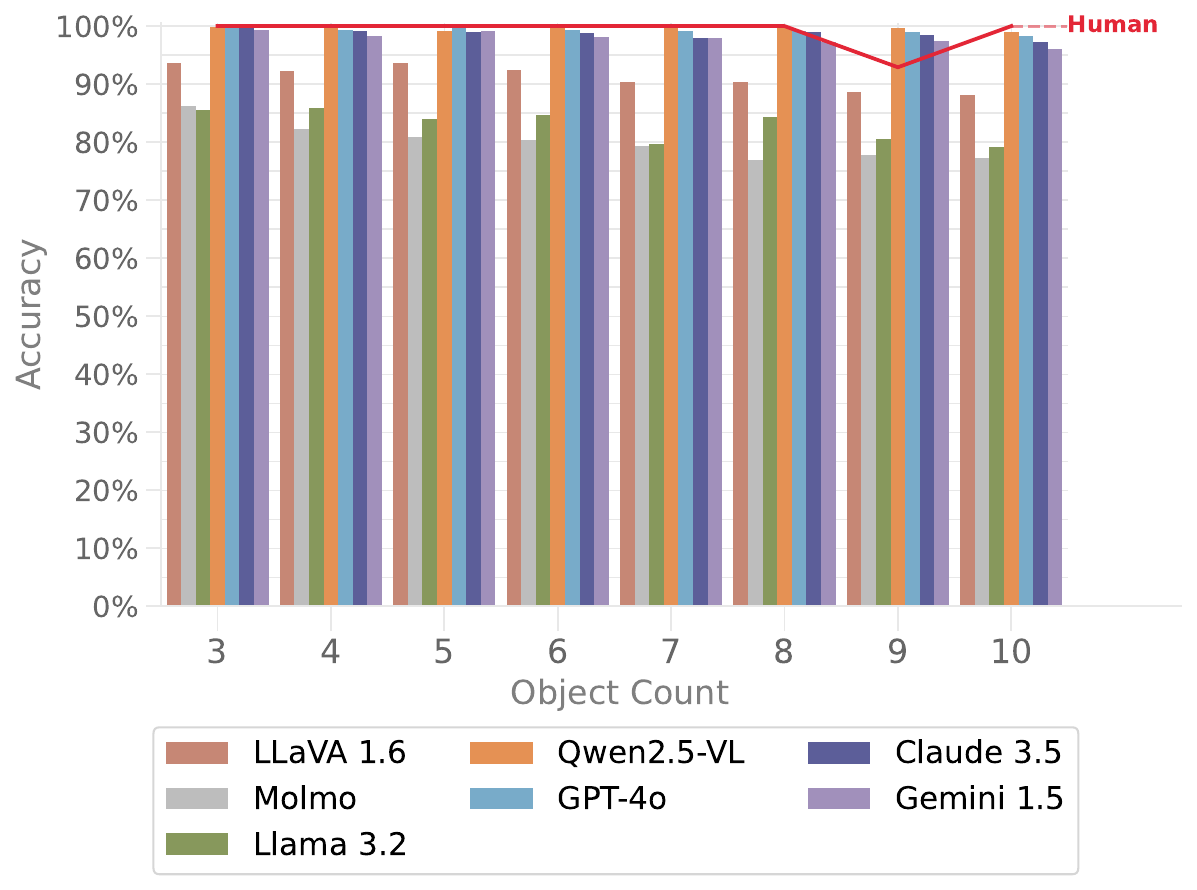}
\caption{Average accuracy of VLMs for text-to-text and image-to-image tasks depending on the number of objects in the scene. VLMs performance keeps stable regardless of the number of objects, just as humans.}
\label{fig:obj_count_performance_unimodal}
\end{figure}

\subsection{Linking attribute identification}
\label{sec:linking-attributes}
In the cross-modal setting, once an object is identified in the pointer modality, it must be cross-referenced with objects in the other modality using attributes that distinguish it from others in the scene. In some cases, a single distinctive attribute is sufficient, while in others it is necessary to combine up to three attributes to uniquely identify the object. We refer to these attributes as linking attributes.

\begin{figure}[t]

\includegraphics[width=\linewidth]{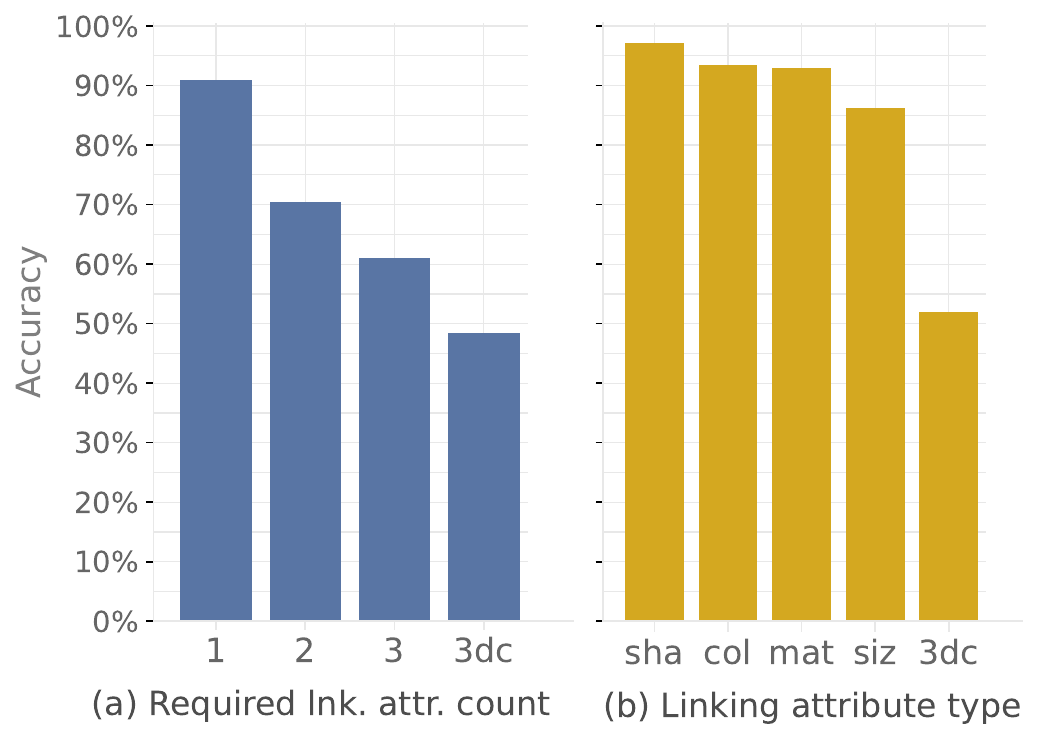}
\caption{(a) Accuracy of Qwen2.5-VL for examples where the correct object must be identified using one, a combination of two, or three visual attributes (i.e., attributes that can be perceived visually). These attributes are considered in addition to 3D coordinates, which could always uniquely identify objects across modalities but models struggle to interpret. This is evident from the accuracy of the examples that rely solely on 3D coordinates (denoted as \textit{3dc}). (b) Accuracy for examples where the correct object can be linked with just one attribute of a single specific type. For example, in the case of material (\textit{mat}), when the pointed object is the only object in the scene made of rubber or metal. \textit{mat}: material, \textit{sha}: \textit{shape}, \textit{col}: color, \textit{siz}: size, \textit{3dc}: 3D coordinates.}
\label{fig:linking_attr}
\end{figure}

In Figure \ref{fig:linking_attr} (a), we analyze how the number of required linking attributes influences the model's performance. We measure Qwen2.5-VL's accuracy, the top-performing open-weight model, on examples where the target entity can be traced with a minimum of one, a combination of two, or three attributes, in addition to the 3D coordinates. Although the 3D coordinates are unique to each entity, using them to correctly identify an object is more complicated than combining other attributes. This can be observed in the last column of the figure, labeled \textit{"3dc"}, where we plot the accuracy for examples where the target object can only be linked using the 3D coordinates. 

Given the significant impact of object count on performance (see Figure \ref{fig:obj_count_performance}), we controlled for this variable by analyzing only scenes with seven objects, as these scenes share an equal distribution of examples requiring one, two, three, or only 3D coordinate linking attributes.

This analysis demonstrates that model performance improves when fewer linking attributes are required, suggesting that the model benefits from a smaller set of distinctive attributes. Furthermore, the model does not appear to simply aggregate all available attributes and directly match the target object. 
In fact, the results indicate that it behaves more like humans, as the accuracy drops noticeably when additional linking attributes are required. If the model used all attributes in a brute-force manner, we would not see this decline in performance.

To shed more light on which type of attribute is most effective as a linking attribute, we measure the model's accuracy for examples that can be linked using a single attribute and separate it by attribute types. These results can be seen in Figure \ref{fig:linking_attr} (b). These attribute types contribute proportionally when combined in pairs or triplets (results for all possible combinations are provided in Appendix \ref{app:all_lnk_attr_performance}). 

These results highlight a clear underperformance of the model in cases where 3D coordinates are the only linking attribute that uniquely identifies the target object. In these cases, the model continues to rely on attribute overlap to make predictions, with a clear preference towards objects with more shared attributes with the target object rather than uniquely relying on the interpretation of 3D coordinates (see Appendix \ref{app:3dc_lnk_attr_overlap}). However, we still see that the model has a partial understanding of 3D coordinate interpretation. In cases where the predicted object matched all attributes except for the 3D coordinates, the model was able to predict the correct object in 63\% of these. 

There are still 14.8\% of cases where the predicted object also contained a mismatch in one or more non-coordinate attributes, indicating that the failure was not uniquely due to misinterpreting spatial information but also involved other attribute-level errors. Even in these cases, the model selected an object with a lower Euclidean distance to the gold object in 70\% of instances, compared to the average distance of objects with the same number of overlapping attributes.

\subsection{Chain-of-Thought Evaluation}
\label{sec:cot}
The main experiments require models to solve the problem in a single step, although humans intuitively approach this task as a series of logical steps. For complex problems, however, VLMs and Large Language Models (LLMs) generally benefit from breaking a task into a sequence of simpler subtasks \cite{wei-etal-2022-cot, chen-etal-2024-measuring}. This process, known as Chain-of-Thought (CoT), helps guide the final output toward the correct solution of a complex problem. 

In order to verify whether poor model performance was due to an inability to handle the task in one step, we conduct a further evaluation of the same task using CoT prompting techniques. Here we prompt the models first to identify the object in the pointer modality that matches the given pointer attribute. Next, they must list this object's attributes, identify those that distinguish it from other objects in the pointer modality (linking attributes), locate the corresponding object in the target modality using these linking attributes, and finally return its target attribute. We conduct this experiment in a two-shot setting (see Appendix \ref{app:cot_prompt} for the complete prompt).

Table \ref{tab:cot} presents the overall results along with their differences compared to the scores without CoT. While CoT significantly improves performance for Molmo and Llama 3.2, it only marginally benefits the commercial models and reduces performance in the LLaVA family. Although some models' performance seems strong overall, especially commercial models, all models still experience performance declines as the number of objects in the scene increases, as shown in the column on the right. This indicates that the main issue still persists and is related to VLMs' inability to link entities across modalities, rather than the multi-step nature of the task. Consequently, improving VLMs for cross-modal entity linking will likely require approaches beyond CoT.

\subsection{Self-Reflection Evaluation}
\label{sec:reasoning_models}

To evaluate whether self-reflective techniques would tackle this problem better,  we evaluated the open-weight self-reflecting visual language model VL-Rethinker-7B \cite{vl-rethinker} using the same 2 shot prompt as models in Table \ref{tab:main_results}. VL-Rethinker-7B achieves 75.7\% and 83.4\% accuracy in the image-to-data and data-to-image settings, respectively. In comparison, Qwen2.5-VL—its base, non-self-reflective counterpart, achieves 75.7 and 84.5 points in the same settings. This suggests that self-reflecting does not provide any added value for cross-modal entity linking.

\begin{table}[t]
\centering
\begin{tabular}{llll}
\toprule
 & Model & All & 10 Obj. \\
\midrule
& Human & 97.9 & 95.8 \\
& Random & 22.0 & 18.6 \\

\midrule

\multirow{5}{*}{\rotatebox{90}{Open}} 

& LLaVA 1.5 & 28.1 \deltaneg{4.5} & 18.5 \deltaneg{6.5} \\
& LLaVA 1.6 & 40.0 \deltaneg{15.2} & 29.7 \deltaneg{8.4} \\
& Molmo & 45.1 \deltapos{25.5} & 31.5 \deltapos{19.0} \\
& Llama 3.2 & 53.7 \deltapos{29.2} & 36.3 \deltapos{24.1} \\
& Qwen2-VL & 72.9 \deltaneg{1.8} & 54.2 \deltapos{0.3} \\
& Qwen2.5-VL & \textbf{78.9} \deltaneg{1.2} & \textbf{62.8} \deltaneg{1.5} \\

\midrule
\multirow{3}{*}{\rotatebox{90}{Closed}} 
& Gemini 1.5 & 72.5 \deltapos{5.1} & 53.2 \deltapos{5.5} \\
& GPT-4o & 82.8 \deltapos{5.0} & 64.6 \deltapos{7.7} \\
& Claude 3.5 & \textbf{86.2} \deltapos{2.9} & \textbf{70.5} \deltapos{4.5} \\

\bottomrule
\end{tabular}
\caption{Results of open and closed VLMs on our task for two-shot CoT prompting. Exact match accuracy is reported for all examples (image-to-text and text-to-image), along with their differences compared to the scores without CoT (originally shown in the Avg. column of Table \ref{tab:main_results}). Results for scenes with 10 objects are also provided.}
\label{tab:cot}
\end{table}

\vspace{-0.2em}
\section{Conclusions}
\label{sec:conclusions}
\vspace{-0.2em}
In this work we demonstrate that VLMs are unable to consistently match the same entity across modalities and retrieve one of its attributes, even guiding their inference with chain-of-thought prompting. To support the evaluation of this fundamental skill, we present MATE, a benchmark designed to assess the proficiency of both current and future models in this area.

Future work could focus on evaluating tasks with greater complexity regarding the pointer and target attributes. One of the main advantages of this benchmark is that it can be easily extended to scenarios involving multiple pointer and target attributes (e.g., \textit{"Identify the name and rotation of the blue sphere"}) and/or a higher number of objects. The last alternative may be especially interesting to evaluate test-time scaling techniques, trying to better align with humans' capability of using more time for more complex scenarios.


Finally, it would be interesting to test models on cross-modal linking in a real-world scenario. One could choose a set of real-world entities (people, cars, etc.) and discretise a set of attributes that uniquely identify the entities (clothing, colour, size, etc). However, the main challenge for such an undertaking is to design the accompanying knowledge base such that you maintain a well-defined scope for the controlled evaluation of cross-modal entity linking.

\section*{Limitations}

MATE is composed of synthetic images, limiting the variety of phenomena that can occur in natural images. We decided to use synthetic images to control for all the important variables of the problem, but we acknowledge that it would be interesting to build a similar dataset with natural images.

The textual part of our benchmark is provided only in English. This decision is based on the fact that current VLMs are mostly trained with English texts. However, as the multilingual capabilities of such models improve, having multilingual versions of our dataset could offer interesting insights. 

Finally, to focus on the cross-modal aligning part, we kept the visual and textual searches simple, using only one pointer attribute. The extension of our benchmark to require combinations of attributes as pointers is straightforward, and could make the task even more difficult.

\section*{Acknowledgments}
This work is partially supported by the Ministry of Science and Innovation of the Spanish Government (AWARE project TED2021-131617B-I00, DeepKnowledge project PID2021-127777OB-C21), project funded by MCIN/AEI/10.13039/501100011033 and by FEDER, the Basque Government (IXA excellence research group IT1570-22), the European Union under Horizon Europe (Project LUMINOUS, grant number 101135724), and the UK Engineering and Physical Sciences Research Council (grant EP/W002876/1).

\bibliography{custom}

\newpage
\appendix

\begin{table}[b!]
    \centering
    \resizebox{\columnwidth}{!}{
    {\rowcolors{2}{black!15!white!40}{black!5!white!40}
    \begin{tabular}{cccccc}
        \toprule
            Nº Attr. & Shape & Material & Color & Size & Acc.  \\
        \midrule
            \cellcolor[HTML]{FFFFFF} 3D Coords & & & & & 48.7 \\
        \midrule
            \cellcolor[HTML]{FFFFFF} & \checkmark & & & & 92.1 \\
            \cellcolor[HTML]{FFFFFF} & & \checkmark & & & 90.1 \\ 
            \cellcolor[HTML]{FFFFFF} & & & \checkmark & & 88.8 \\ 
            \cellcolor[HTML]{FFFFFF} \multirow{-4}{*}{1} & & & & \checkmark & 76.9 \\
        \midrule
            \cellcolor[HTML]{FFFFFF} & \checkmark & \checkmark & & & 70.7 \\
            \cellcolor[HTML]{FFFFFF} & \checkmark & & & \checkmark & 54.8 \\
            \cellcolor[HTML]{FFFFFF} & & \checkmark & \checkmark & & 68.5 \\
            \cellcolor[HTML]{FFFFFF} & & \checkmark & & \checkmark & 58.0 \\
            \cellcolor[HTML]{FFFFFF} \multirow{-5}{*}{2} & & & \checkmark & \checkmark & 52.2 \\
        \midrule
            \cellcolor[HTML]{FFFFFF} & \checkmark & \checkmark & & \checkmark & 44.2 \\
            \cellcolor[HTML]{FFFFFF} \multirow{-2}{*}{3} & & \checkmark & \checkmark & \checkmark & 57.1 \\
        \bottomrule
    \end{tabular}
    }
    }
    \captionof{table}{Average performance in instances where different linking attribute types are necessary to align objects across modalities, ordered by the number of attributes.} 
    \label{tab:link_attr_all}
\end{table}

\begin{table*}[!]
\centering
\begin{tabular}{lll}
\toprule
Attr. name & Type & Values \\
\midrule
Material & Discrete & \{metal, rubber\} \\
Shape & Discrete & \{sphere, cube, cylinder, cone\} \\
Color & Discrete & \{blue, green, red, gray, cyan, brown, yellow, purple\} \\
Name & Discrete & \{Object\_i\} where $i \in [0, 9]$ \\
Size & Discrete & \{0.35, 0.351, 0.7, 0.701\} \\
Rotation & Continuous & [0, 360] \\
3D coordinates & Continuous & $[x, y, z]$ where $(x, y, z) \in$ ($-\infty, +\infty$) \\
\bottomrule
\end{tabular}
\caption{The complete list of attributes we use in the MATE dataset.}
\label{tab:attr-descr}
\end{table*}

\section{Benchmark Attributes}
\label{app:attributes}

Table \ref{tab:attr-descr} contains the attribute list used in our benchmark, alongside the range of values each attribute can have.

\section{MATE Benchmark Dataset Format}
\label{app:dataset_format}

Each instance of MATE contains key metadata for the definition of the scene, input prompt and key attributes needed to solve the task at hand. Table \ref{tab:mate_data} describes these metadata.

\begin{table*}[!]
    \centering
    \begin{tabular}{lcp{9cm}}
        \toprule
            Name & \multicolumn{1}{l}{Type}  & Description  \\
        \midrule
            \emph{example\_id} & str & Uniquely identifiable id of this instance. \\
            \emph{task} & str & Task of this example: img2img, txt2txt, img2txt or txt2img. \\
            \emph{input\_str} & str & Input prompt that includes task definition, few-shot examples (when applicable) and question to be answered. \\
            \emph{gold\_reference} & str & The expected answer by the model. \\
        \midrule
            \emph{image} & str & Filename of the scene's image. \\
            \emph{scene} & JSON & Serialized scene containing all objects and their attributes. \\
            \emph{scene\_format} & str & Format used to represent the scene: JSON (by default), YAML, XML or TXT. \\ 
            \emph{object\_count} & int & Number of objects in the scene (from 3 to 10). \\
        \midrule
            \emph{pointer\_attribute} & JSON & Pointer attribute and its value. \\
            \emph{target\_attribute} & JSON & Target attribute and its value. \\
            \emph{key\_attributes} & list & Combination of attributes that uniquely identify the target object (not including the pointer and target attributes). \\ 
            \emph{few\_shot\_attributes} & list & Pointer, key and target attributes of few-shot examples. \\
        \bottomrule
    \end{tabular}
    \caption{The complete list of metadata we use in the MATE dataset with detailed descriptions and data types.}
    \label{tab:mate_data}
\end{table*}

\section{Participants in Human Evaluation}
\label{app:participants}

The participants in the human evaluation ranged from 28 to 44 years old and all had university degrees.

\section{All Evaluated Model Results}
\label{app:all_results}

We evaluated a total of 16 models in our benchmark, MATE, from several families of VLMs. Table \ref{tab:all_eval_models} summarizes the results we obtained with them, both in unimodal and crossmodal settings. It is an extended version of Tables \ref{tab:main_results} and \ref{tab:unimodal_results}.

\begin{table*}[th]
\centering
\begin{tabular}{lrrrrrr}
    \toprule
        \multirow{2}{*}{\vspace{-0.2cm}Model} & \multicolumn{3}{c}{{Unimodal}} & \multicolumn{3}{c}{{Crossmodal}} \\
     \cmidrule(lr){2-4}  \cmidrule(lr){5-7} 
        & Img2Img & Txt2Txt & Avg. & Img2Txt & Txt2Img & Avg. \\
    \midrule
        Human & 100.0 & 99.0 & 99.5 & 97.9 & 97.9 & 97.9 \\
        Random & 18.7 & 25.3 & 22.0 & 25.4 & 18.5 & 22.0 \\ 
    \midrule
llava-1.5-7b & 45.7 & 60.7 & 53.2 & 18.2 & 27.5 & 22.9 \\
llava-1.5-13b & 55.3 & 88.6 & 71.9 & 29.3 & 35.7 & 32.5 \\
llava-v1.6-mistral-7b & 77.3 & 92.4 & 84.8 & 40.6 & 37.1 & 38.9 \\
llava-v1.6-vicuna-7b & 58.6 & 75.9 & 67.2 & 29.0 & 32.2 & 30.6 \\
llava-v1.6-vicuna-13b & 73.9 & 92.5 & 83.2 & 35.0 & 38.4 & 36.7 \\
llava-v1.6-yi-34b & 84.4 & 98.0 & 91.2 & 48.7 & 61.6 & 55.2 \\
llama3-llava-next-8b & 80.4 & 98.2 & 89.3 & 44.7 & 50.5 & 47.6 \\
MolmoE-1B-0924 & 10.5 & 19.0 & 14.8 & 17.8 & 21.8 & 19.8 \\
Molmo-7B-O-0924 & 88.3 & 71.9 & 80.1 & 32.3 & 31.4 & 31.8 \\
Molmo-7B-D-0924 & 86.9 & 54.7 & 70.8 & 18.1 & 20.9 & 19.5 \\
Llama-3.2-11B-Vision & 68.9 & 97.1 & 83.0 & 37.4 & 11.4 & 24.4 \\
Qwen2-VL-2B-Instruct & 99.1 & 81.4 & 90.2 & 44.1 & 44.8 & 44.5 \\
Qwen2-VL-7B-Instruct & \textbf{99.7} & 98.1 & 98.9 & 72.1 & 77.2 & 74.7 \\ 
Qwen2.5-VL-7B-Instruct & \textbf{99.7} & \textbf{99.4} & \textbf{99.5} &\textbf{ 75.7} & \textbf{84.5} & \textbf{80.1} \\ 
    \midrule
        gemini-1.5-flash & 95.9 & \textbf{100.0} & \textbf{95.9} & 63.2 & 71.2 & 67.2 \\
        gpt-4o-2024-11-20 & \textbf{98.4} & \textbf{100.0} & \textbf{99.2} & 76.4 & 79.1 & 77.8 \\
        claude-3-5-sonnet-20241022 & 97.3 & \textbf{100.0} & 98.7 & \textbf{80.9} & \textbf{85.7} & \textbf{83.3} \\
    \bottomrule
\end{tabular}
\caption{Results of open and closed VLMs in our task. All results are obtained using 2-shot prompting. Exact match accuracy is provided for image-to-image (Img2Img column), text-to-text (Txt2Txt column), image-to-text (Img2Txt column), and text-to-image (Txt2Img column) configurations. Human and random accuracies are shown as reference.}
\label{tab:all_eval_models}
\end{table*}

\section{Predicted Object Attribute Overlapping in 3D Coordinate-Only Linking Attribute Cases}
\label{app:3dc_lnk_attr_overlap}

This section analyzes cases where 3D coordinates are the only linking attribute that uniquely identifies the target object. In Figure \ref{fig:obj_count_performance_unimodal}, we report how many attributes of Qwen2.5-VL’s predicted objects match the target's.

\begin{figure}[]
\includegraphics[width=\linewidth]{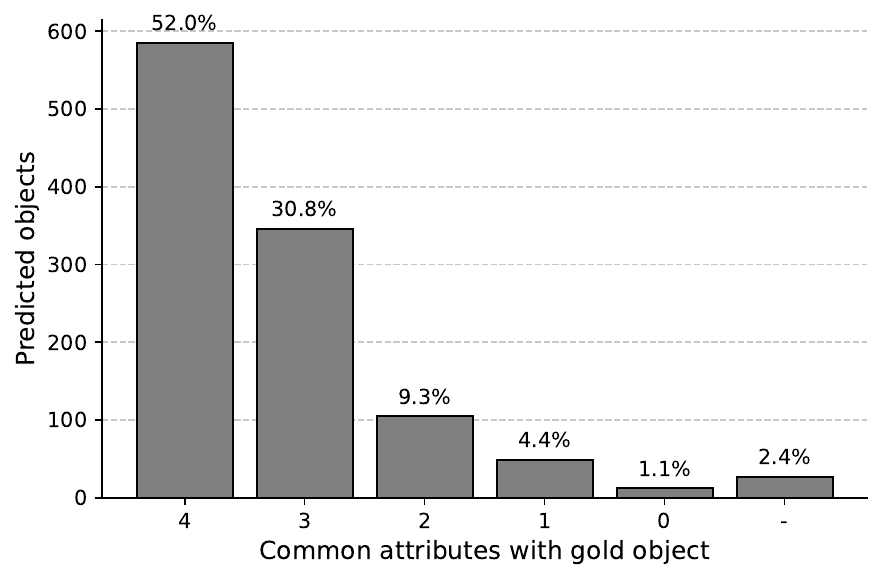}
\caption{Number of matching attributes between Qwen2.5-VL’s predicted object and the target object in cases where 3D coordinates are the only uniquely identifying linking attribute. A value of '4' indicates that the model selected the correct object (i.e., all attributes matched), while '–' indicates predictions that could not be linked to any existing object.}
\label{fig:obj_count_performance_unimodal}
\end{figure}

\section{Zero- and Few-shot Experiments}
\label{app:shot_experiments}
In Table \ref{tab:all_eval_models} we show the results of all the VLMs for zero-, one- and two-shot prompting. Image-to-text and text-to-image exact match accuracies are provided for every variant.

\begin{table*}[th]
\centering
\begin{tabular}{lrrrrrr}
    \toprule
     Model & \multicolumn{3}{c}{{Image2Text}} & \multicolumn{3}{c}{{Text2Image}} \\
     \cmidrule(lr){2-4}  \cmidrule(lr){5-7} 
     & 0 & 1 & 2 & 0 & 1 & 2 \\
    \midrule
llava-1.5-7b & 19.4 & 16.3 & 18.2 & 7.6 & 27.6 & 27.5 \\
llava-1.5-13b & 20.4 & 27.2 & 29.3 & 17.1 & 33.4 & 35.7 \\
llava-v1.6-mistral-7b & 36.7 & 38.1 & 40.6 & 34.6 & 36.3 & 37.1 \\
llava-v1.6-vicuna-7b & 26.3 & 28.0 & 29.0 & 11.4 & 29.6 & 32.2 \\
llava-v1.6-vicuna-13b & 26.1 & 32.7 & 35.0 & 7.7 & 33.4 & 38.4 \\
llava-v1.6-yi-34b & 43.2 & 46.8 & 48.7 & 54.5 & 55.9 & 61.6 \\
Llama3-llava-next-8b & 40.6 & 43.5 & 44.7 & 40.5 & 49.1 & 50.5 \\
MolmoE-1B-0924 & 3.3 & 19.8 & 17.8 & 5.5 & 17.5 & 21.8 \\
Molmo-7B-O-0924 & 31.4 & 24.9 & 32.3 & 19.1 & 33.8 & 31.4 \\
Molmo-7B-D-0924 & 12.2 & 23.5 & 18.1 & 18.1 & 23.9 & 20.9 \\
Llama-3.2-11B-Vision & 27.1 & 36.1 & 37.4 & 4.0 & 12.1 & 11.4 \\
Qwen2-VL-2B-Instruct & 26.9 & 39.4 & 44.1 & 42.8 & 34.1 & 44.8 \\
Qwen2-VL-7B-Instruct & 68.8 & 72.4 & 72.1 & 77.4 & 77.9 & 77.2 \\ 
Qwen2.5-VL-7B-Instruct & \textbf{72.0} & \textbf{74.8} & \textbf{75.7} & \textbf{83.6} & \textbf{82.9} & \textbf{84.5} \\
    \midrule
gemini-1.5-flash & 59.1 & 61.9 & 63.2 & 63.6 & 68.9 & 71.2 \\
gpt-4o-2024-11-20 & 74.8 & 75.5 & 76.4 & 78.8 & 80.2 & 79.1 \\
claude-3-5-sonnet-20241022 & \textbf{77.8} & \textbf{79.9} & \textbf{80.9} & \textbf{83.5} & \textbf{84.9} & \textbf{85.7} \\
    \bottomrule
\end{tabular}
\caption{Results of open and closed VLMs in our task with zero, 1, and 2 shot prompting. Exact match accuracy is provided for image-to-text (Image2Text column) and text-to-image (Text2Image column) configurations.}
\label{tab:all_eval_models}
\end{table*}

\section{Linking Attribute Performance}
\label{app:all_lnk_attr_performance}
Linking attributes have to be used to identify the pointed object in the target modality. Sometimes, one attribute is enough for this process, but other times combinations of two or three attributes are required. In Table \ref{tab:link_attr_all}, we show the average accuracy obtained for different combinations of attributes. Notice that 3D coordinates are treated apart since 3D coordinates are always unique.

\section{Evaluation prompts}
In this appendix, we provide the specific prompts used to evaluate VLMs in this work.

\subsection{UniModal Evaluation Prompts}
\label{app:single_prompt}
\begin{figure*}[p]
\centering
\includegraphics[scale=.85]{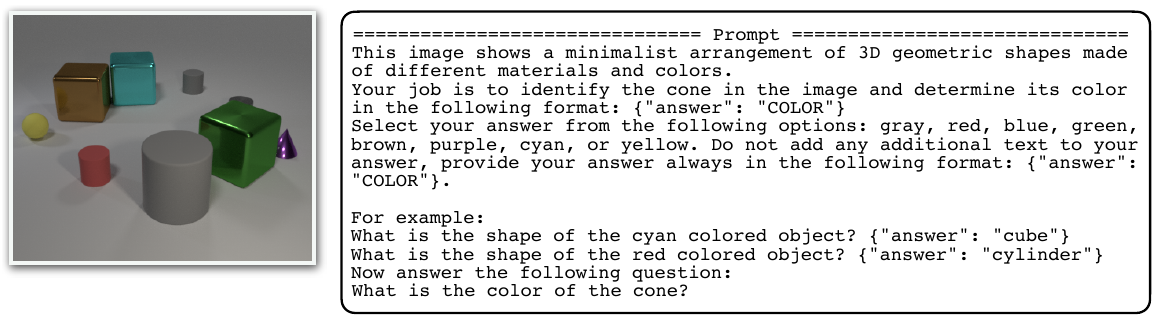}
\caption{Prompt used for image-to-image.}
\label{fig:prompt_img2img}
\end{figure*}

\begin{figure*}[p]
\centering
\includegraphics[scale=.85]{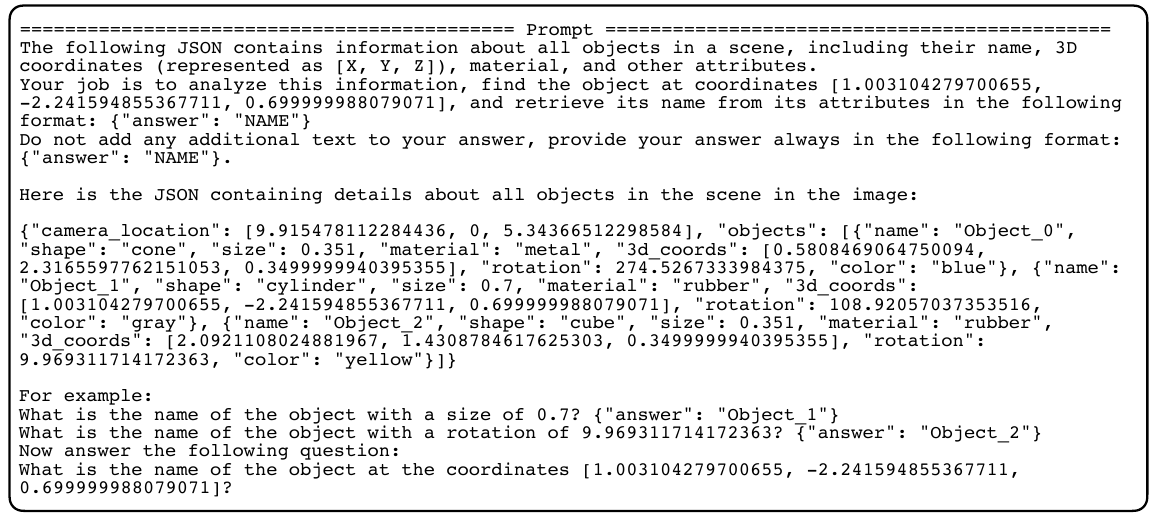}
\caption{Prompt used for text-to-text.}
\label{fig:prompt_txt2txt}
\end{figure*}

Unimodal experiments are divided between text-to-text and image-to-image tasks. Figure \ref{fig:prompt_txt2txt} shows the prompt we use for text-to-text tasks, whereas Figure \ref{fig:prompt_img2img} shows the prompt for image-to-image tasks.


\subsection{Cross-modal Evaluation Prompts}
\label{app:croxss_prompt}
\begin{figure*}[p]
\centering
\includegraphics[scale=.85]{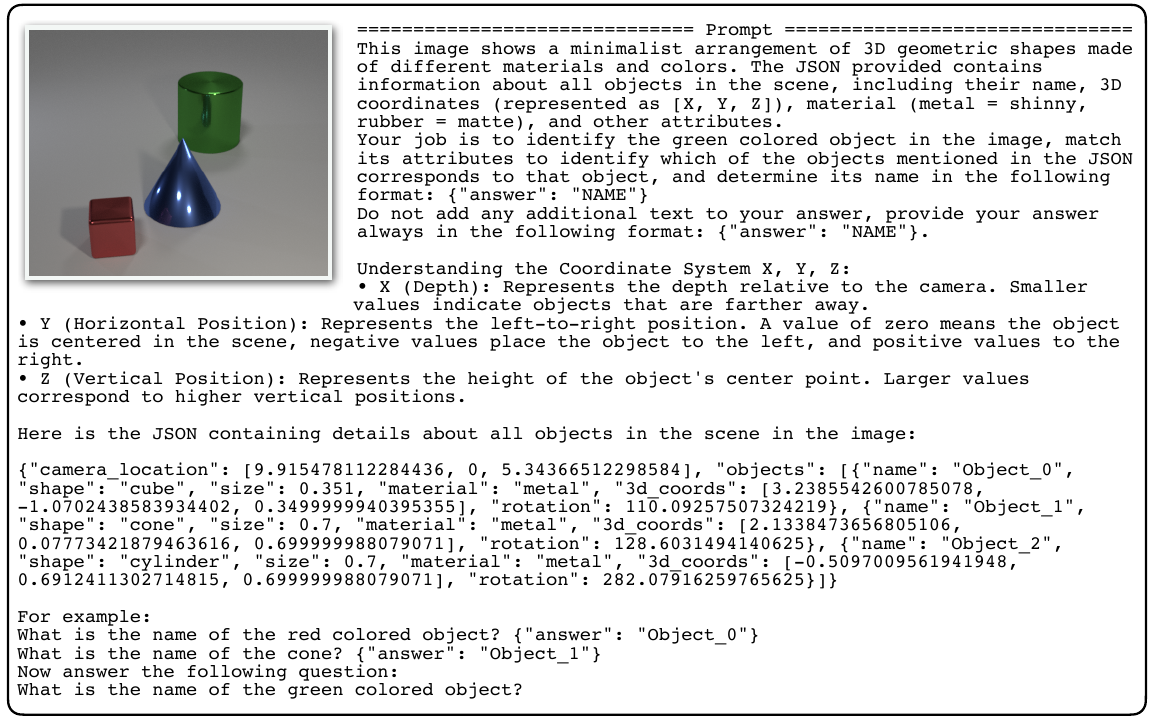}
\caption{Prompt used for image-to-text.}
\label{fig:prompt_img2txt}
\end{figure*}

\begin{figure*}[p]
\centering
\includegraphics[scale=.85]{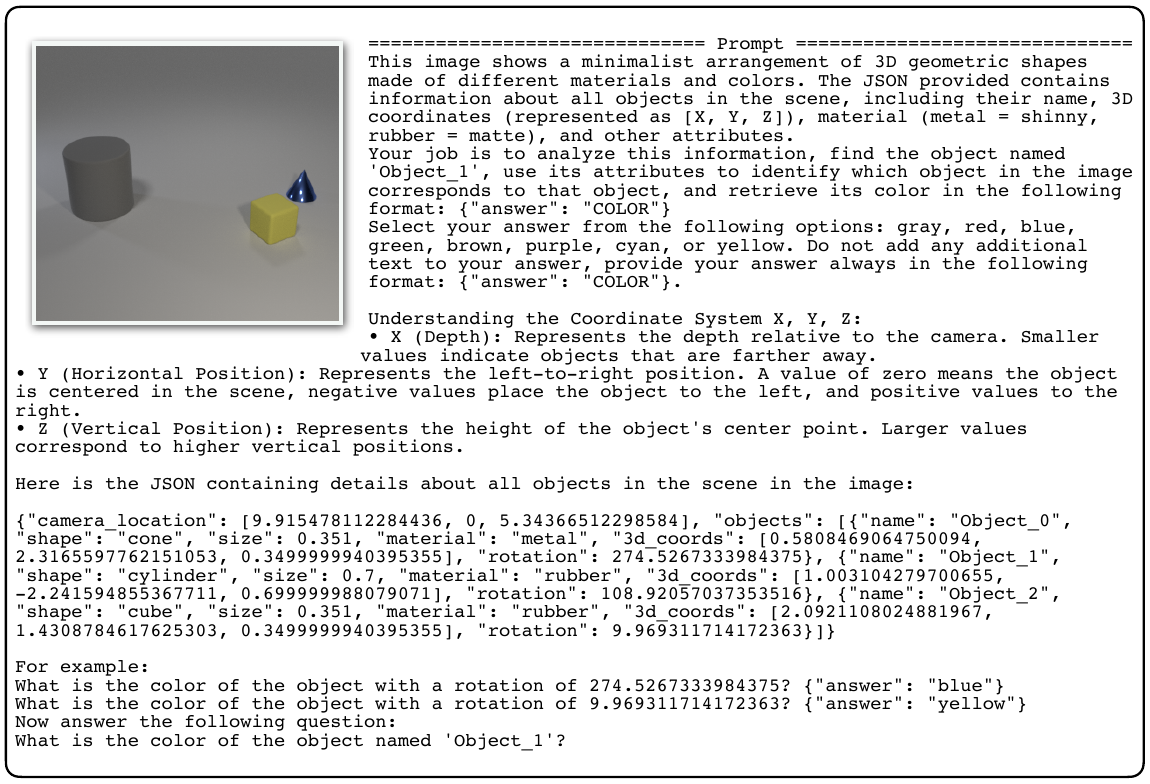}
\caption{Prompt used for text-to-image.}
\label{fig:prompt_txt2img}
\end{figure*}

For cross-modal evaluation we also use two prompts, depending on the task variant (image-to-text or text-to-image). Figure \ref{fig:prompt_img2txt} is used for image-to-text and Figure \ref{fig:prompt_txt2img} for text-to-image.

\subsection{Chain-of-Thought}
\label{app:cot_prompt}
\begin{figure*}[p]
\centering
\includegraphics[scale=.85]{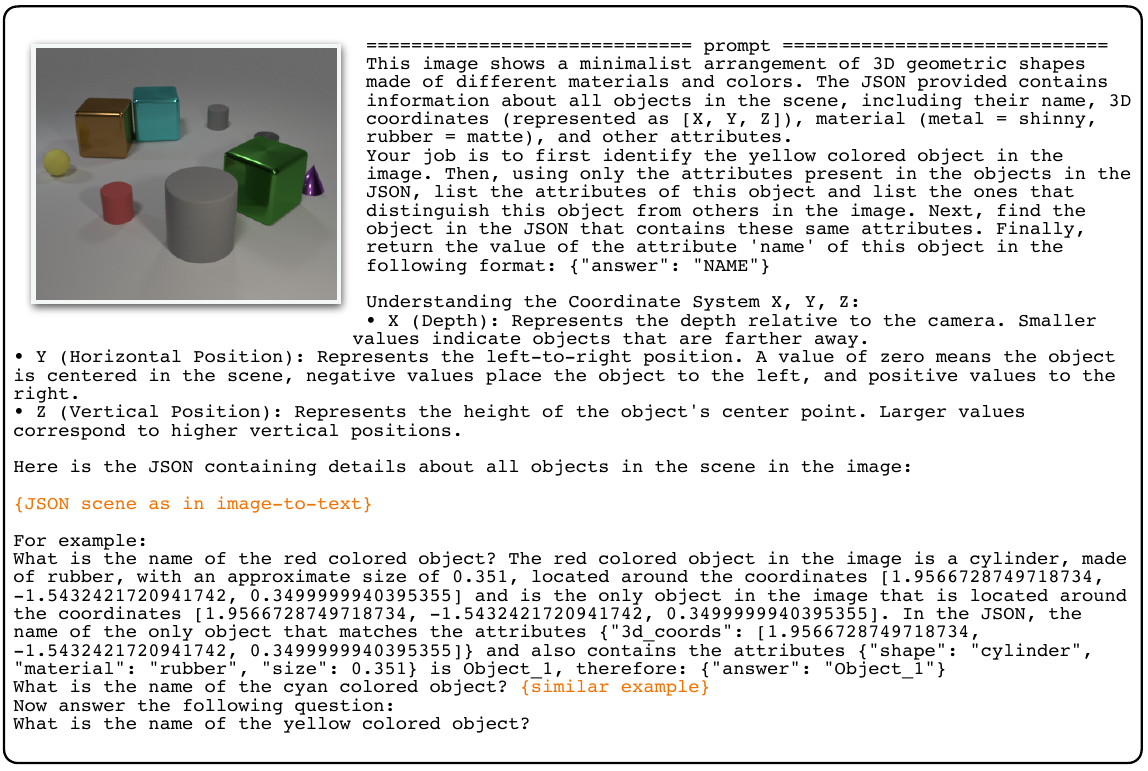}
\caption{Prompt used for image-to-text with Chain-of-Thought.}
\label{fig:prompt_img2txt_cot}
\end{figure*}

\begin{figure*}[t!]
\centering
\includegraphics[scale=.85]{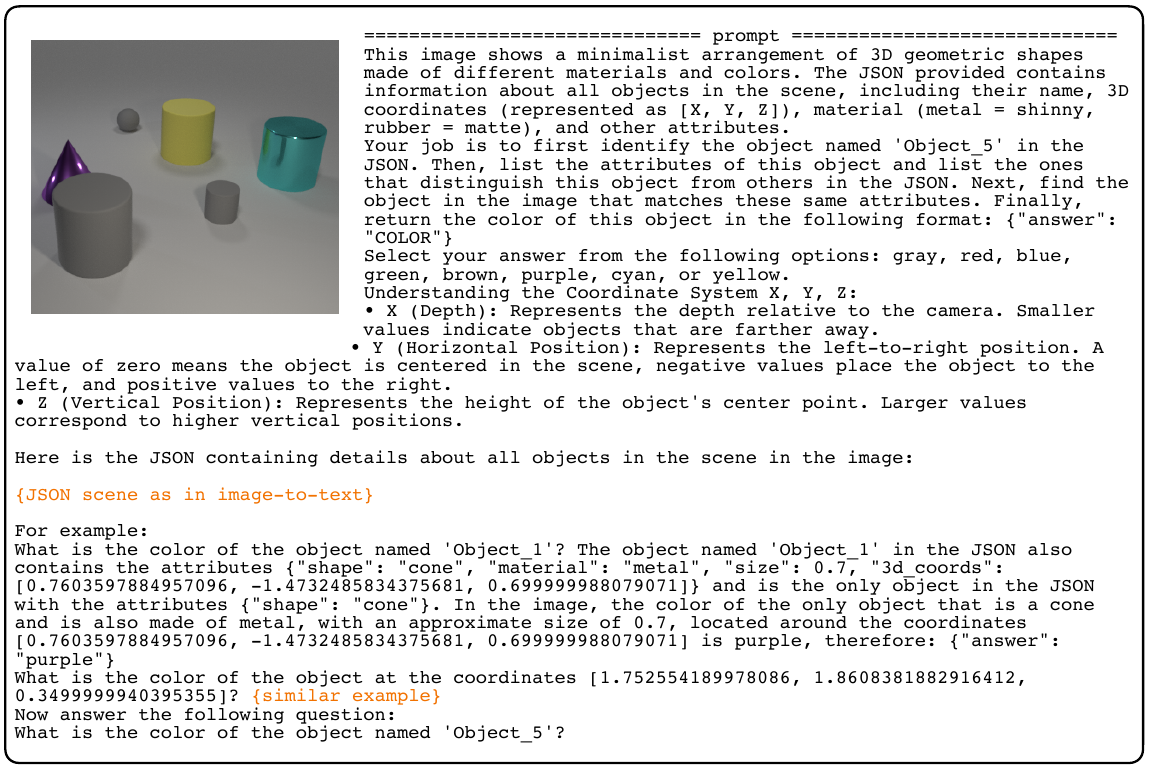}
\caption{Prompt used for text-to-image with Chain-of-Thought.}
\label{fig:prompt_txt2img_cot}
\end{figure*}

The prompts used for the chain-of-thought experiments are shown in Figure \ref{fig:prompt_img2txt_cot} (for the image-to-text task) and \ref{fig:prompt_txt2img_cot} (for text-to-image). 

\newpage

\end{document}